\documentclass[runningheads]{llncs}

\usepackage[T1]{fontenc}
\usepackage{graphicx}

\usepackage{amsmath}
\usepackage{amssymb}
\usepackage{textcomp}
\usepackage{multirow}
\usepackage{array}
\usepackage{algorithm}
\usepackage{algpseudocode}
\usepackage{booktabs}
\usepackage{verbatim}
\usepackage{soul}
\usepackage[caption=false]{subfig}

\usepackage[bookmarks,hidelinks]{hyperref}

\makeatletter
\def\ps@headings{\let\@mkboth\@gobbletwo
  \let\@oddfoot\@empty\let\@evenfoot\@empty
  \def\@evenhead{\normalfont\small\hspace{\headlineindent}%
                 \leftmark\hfil}
  \def\@oddhead{\normalfont\small\hfil\rightmark\hspace{\headlineindent}}
  \def\chaptermark##1{}%
  \def\sectionmark##1{}%
  \def\subsectionmark##1{}}
\makeatother
\usepackage{color}

\definecolor{mainkeycolor}{RGB}{0, 123, 255}
\definecolor{attrcolor}{RGB}{232, 62, 140}
\definecolor{NewColor}{rgb}{0.2,0,0.5}

\algnewcommand\algorithmicinput{\textbf{Input:}}
\algnewcommand\Input{\item[\algorithmicinput]}
\algnewcommand\algorithmicoutput{\textbf{Output:}}
\algnewcommand\Output{\item[\algorithmicoutput]}

\begin{document}

\title{PhenSPINE: A Standardized Benchmark for Spine Pathology Diagnosis}
\titlerunning{PhenSPINE}
%
\author{Duong Ngoc Vu\inst{1} \and
Hai Son Nguyen\inst{1} \and
Trong-Nghia Nguyen\inst{1} \and
Bien Tran Van\inst{3,4} \and
Trang Mai Xuan\inst{2}\thanks{Corresponding author} \and Huan Vu\inst{1} \and
Thien Van Luong\inst{1}}
\authorrunning{D. N. Vu \textit{et al.}}
%

\institute{Business AI Lab, College of Technology, National Economics University, Vietnam\\
\and
A2I Lab, Phenikaa School of Computing, Phenikaa University, Hanoi, Vietnam\\
\and
Medical Imaging \& Radiological Technology Department, Faculty of Medical Technology, Phenikaa School of Medicine \& Pharmacy, Phenikaa University, Vietnam\\
\and
Radiology and Functional Exploration Center, Phenikaa University Hospital, Vietnam
\\
\email{duongvn.bai@st.neu.edu.vn, hains24206@gmail.com, nghiant@neu.edu.vn, bien.tranvan@phenikaa-uni.edu.vn, trang.maixuan@phenikaa-uni.edu.vn, \{huanv, thienlv\}@neu.edu.vn}
}

\maketitle              
\begin{abstract}
The accurate diagnosis of spinal pathologies depends heavily on radiological interpretation, yet automated systems are hindered by the lack of diverse, high-quality benchmarks. In this study, we present PhenSPINE, a Magnetic Resonance Imaging dataset comprising 16,813 images from 250 patients, curated to facilitate advanced deep learning research. We propose a robust diagnostic benchmark that integrates state-of-the-art convolutional backbones with a Positional Encoding mechanism to explicitly model the anatomical context of intervertebral discs. Evaluating across four standard MRI sequences, our experiments demonstrate that the Sagittal T2-weighted sequence offers the most robust diagnostic value, achieving a superior Macro F1-score of 50.31\%. We find that multi-sequence fusion strategies yield inferior performance compared to this single-sequence baseline, as the images across sequences in our dataset are significantly compromised by noise interference from surrounding anatomical regions. This work establishes a robust baseline and offers critical insights into sequence selection for spine analysis.

\end{abstract}

\section{Introduction}
The human spine is a sophisticated anatomical axis, integrating osseous, ligamentous, and muscular components to ensure biomechanical stability and protect the central nervous system. Despite its structural robustness, the spinal column is susceptible to a heterogeneous array of pathologies. Currently, definitive diagnosis relies on a synthesis of clinical assessment and radiological imaging, often supplemented by invasive tissue sampling. However, biopsies are frequently hindered by sampling inaccuracies and the inherent risks of surgical complications \cite{rolscamp1997complications,nasser2010complications}. The temporal lag associated with histopathological turnaround can be detrimental to the patient’s oncological and functional prognosis \cite{alshieban2015reducing}.

Transitioning towards a purely Magnetic Resonance Imaging (MRI)-based diagnostic paradigm offers a significant opportunity to accelerate clinical workflows, mitigate procedural risks, and enhance global accessibility to specialized care. In recent years, the medical field has witnessed a transformative shift toward non-invasive, data-driven methodologies, with Deep Learning emerging as the predominant architecture for complex medical image analysis due to its superior hierarchical feature extraction capabilities \cite{esteva2019guide,grossman2021differentiating}. 

To bridge the gap between algorithmic potential and clinical utility, this paper provides several key contributions. (1) We introduce PhenSPINE: a comprehensive dataset specifically curated for spinal pathology diagnosis. (2) We develop and evaluate benchmark for four distinct MRI sequences, establishing a performance baseline across diverse imaging sequences. (3) We propose a Positional Encoding mechanism to explicitly model the anatomical location of intervertebral discs, significantly improving diagnostic performance by leveraging spatial context. (4) Our empirical analysis identifies the Sagittal T2-weighted (SAG-T2) sequence as the most informative sequence, yielding the highest diagnostic precision. (5) We demonstrate that integrating all four MRI sequences leads to performance degradation compared to the best-performing single-sequence model, highlighting noise in spinal image fusion in our dataset.

The remainder of this paper is organized as follows. Section~\ref{sec:related_work} reviews related work. Section~\ref{sec:materials_and_method} details the PhenSPINE dataset and the benchmark. Section~\ref{sec:experiments} analyzes the experimental results. Finally, Section~\ref{sec:conclusion} concludes the paper.

\section{Related Work}
\label{sec:related_work}
\subsection{Spine Medical Imaging Datasets.}
Access to high-quality, annotated datasets is a prerequisite for developing robust computer-aided diagnosis systems. The UW-Spine dataset \cite{glocker2013vertebrae}, comprised of CT scans from 125 patients, served as a foundational resource for vertebra localization, particularly in pathological cases involving scoliosis and metal implants. Following this, the Large Scale Vertebrae Segmentation Challenge datasets \cite{sekuboyina2021verse} established a gold standard for volumetric spine segmentation. These datasets provide diverse CT scans with voxel-level annotations and centroid coordinates, addressing challenges in diverse field-of-views and anatomical variations. In the domain of Magnetic Resonance Imaging, the Lumbar Spine MRI Dataset \cite{van2024lumbar} has been widely utilized for intervertebral disc classification and Pfirrmann grading. Most recently, the RSNA 2024 Lumbar Spine Degenerative Classification competition \cite{richards2026rsna} introduced a large-scale, multi-center MRI dataset labeled by expert neuroradiologists. Unlike previous datasets restricted to single pathologies, RSNA 2024 necessitates the simultaneous classification of multiple degenerative conditions-lumbar disc herniation, spinal canal stenosis, and neural foraminal narrowing-across five lumbar levels. In this work, we introduce PhenSPINE, a dataset designed to evaluate multi-sequence models for spinal diagnostics.

\subsection{Deep Learning Approaches for Spine Pathology Diagnosis.}
Early efforts in spine analysis, particularly on datasets like UW-Spine, relied heavily on traditional machine learning techniques. Glocker \textit{et al.} \cite{glocker2013vertebrae} employed regression forests for vertebrae localization, effectively handling pathological deformations without deep neural networks. However, the paradigm shifted with the advent of large-scale benchmarks. In segmentation tasks such as the VerSe challenge, a new performance standard was defined in \cite{payer2020coarse} through the application of U-Net \cite{ronneberger2015u} and its variants within a coarse-to-fine framework. This approach first regresses a global heatmap to localize the spine and subsequently segments individual vertebrae with high precision. In the domain of pathology diagnosis on MRI, deep learning has replaced manual feature engineering. Natalia \textit{et al.} \cite{natalia2024lumbar} utilized these architectures to automate the classification of intervertebral discs and Pfirrmann grading. We are the first to apply positional encoding combined with a multi-sequence fusion approach for spine pathology diagnosis.

In this work, we propose a diagnostic benchmark that integrates Positional Encoding to explicitly model the anatomical context of intervertebral discs. The specific details of these contributions, including the dataset characteristics and the proposed methodology, will be presented in the section below.

\section{Materials and method}
\label{sec:materials_and_method}
\subsection{PhenSPINE Dataset}
\label{subsec:phenspine_benchmark}
\subsubsection{Data Collection.}
\begin{figure}[ht]
    \centering
    \includegraphics[width=0.7\linewidth, trim=1.1cm 0 0 0, clip]{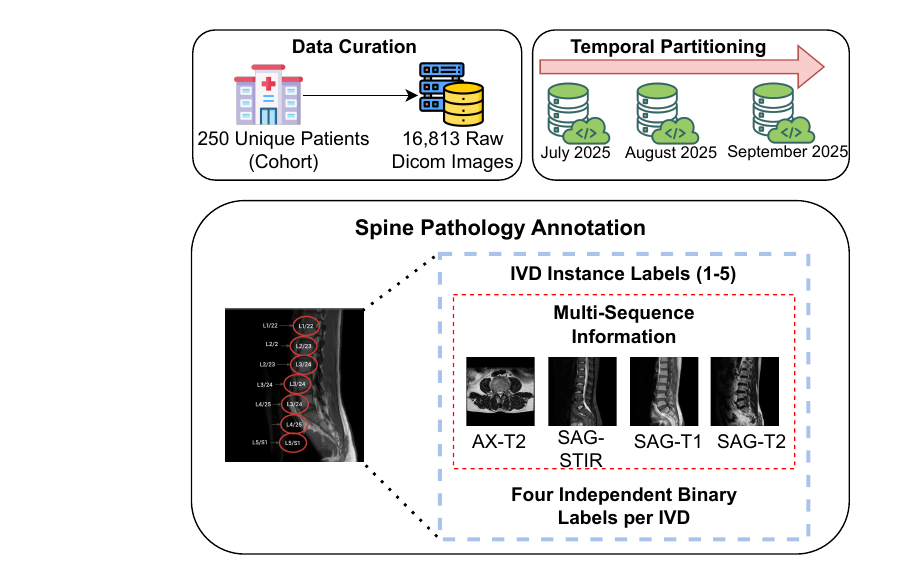}
    \caption{Overview of the PhenSPINE data collection process.}
    \label{fig:data_collection}
\end{figure}
A total of 16,813 raw DICOM images representing a cohort of 250 unique patients were curated at PHENIKAAMEC hospital and annotated by expert physicians, as illustrated in Fig.~\ref{fig:data_collection}. To ensure temporal diversity and mitigate potential bias associated with longitudinal variations in imaging equipment or protocols, the benchmark was partitioned into three chronological subsets collected throughout 2025, as summarized in Table~\ref{tab:data_summary}. Regarding spine pathology annotations, each IVD instance is specified by four independent binary labels, its anatomical location within the lumbar spine (IVD labels 1 to 5), and the integrated multi-sequence information comprising Sagittal T2-weighted (SAG-T2), Sagittal T1-weighted (SAG-T1), Axial T2-weighted (AX-T2), and Sagittal STIR (SAG-STIR).

\begin{table}[htbp]
\centering
\caption{Quantitative summary of the PhenikaaMed MRI subsets}
\label{tab:data_summary}
\begin{tabular}{lccc}
\hline
\textbf{Subset} & \textbf{Patients} & \textbf{DICOM Files} & \textbf{Collection Period} \\ \hline
T27.7.25        & $51$              & $3,458$              & July 2025                \\
T8              & $70$              & $4,576$              & August 2025              \\
T9              & $129$             & $8,779$              & September 2025             \\ \hline
\textbf{Total}  & $\mathbf{250}$    & $\mathbf{16,813}$    & \textbf{--}              \\ \hline
\end{tabular}
\end{table}

\subsubsection{Data Distribution.}
\begin{figure}[htbp]
    \centering
    \subfloat[Herniation]{
        \includegraphics[width=0.23\linewidth]{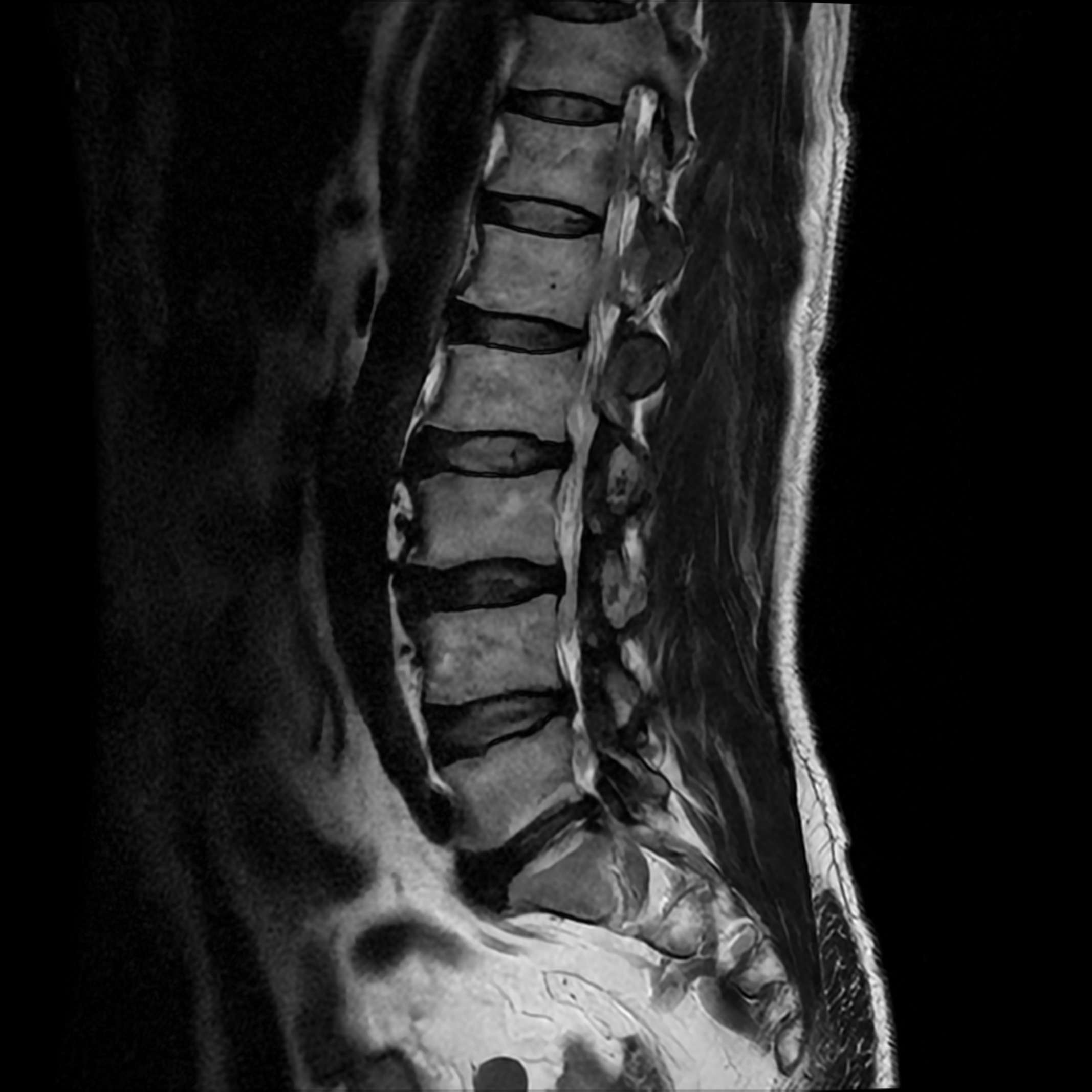}
    }
    \hfill
    \subfloat[Bulging]{
        \includegraphics[width=0.23\linewidth]{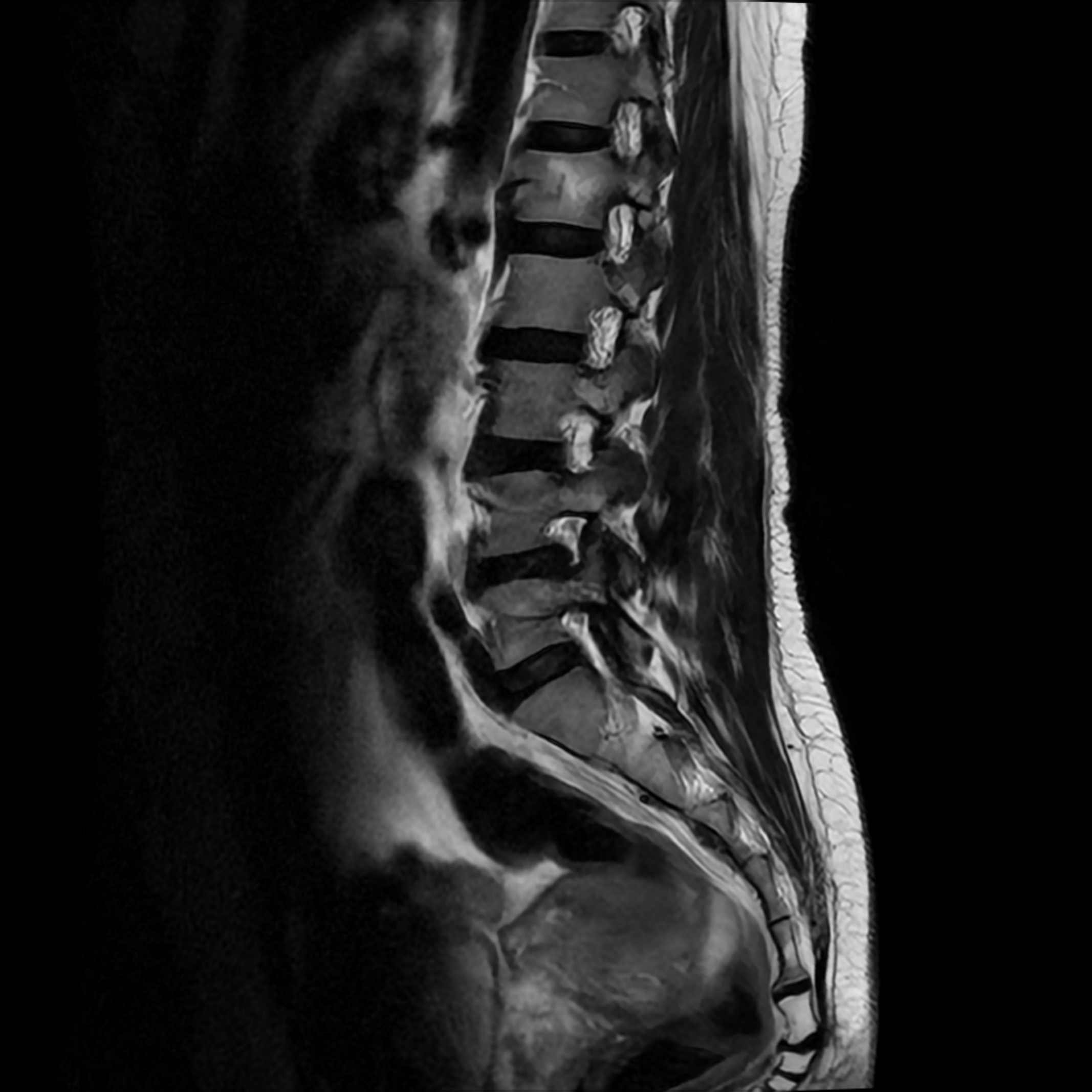}
    }
    \hfill
    \subfloat[Spondylolisthesis]{
        \includegraphics[width=0.23\linewidth]{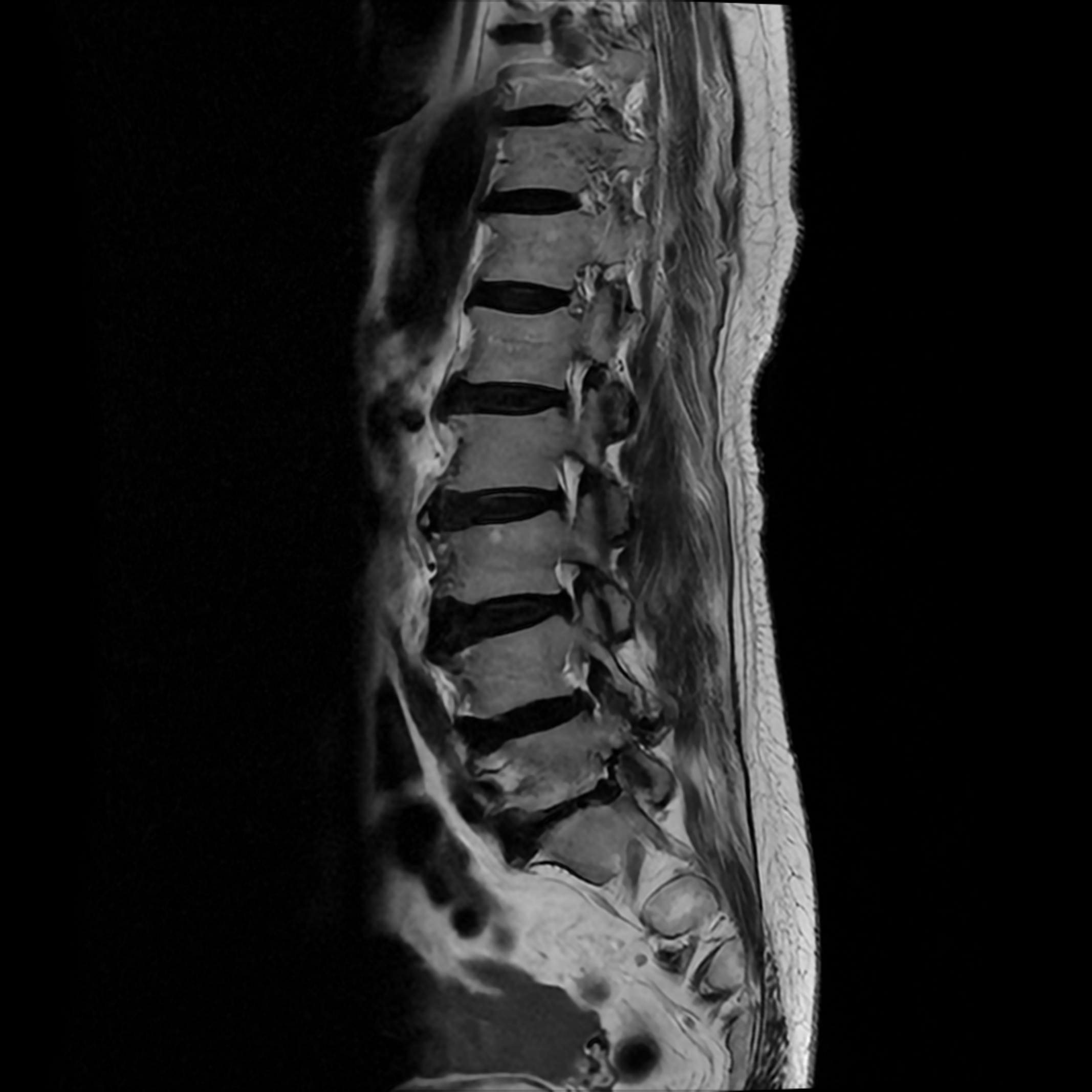}
    }
    \hfill
    \subfloat[Narrowing]{
        \includegraphics[width=0.23\linewidth]{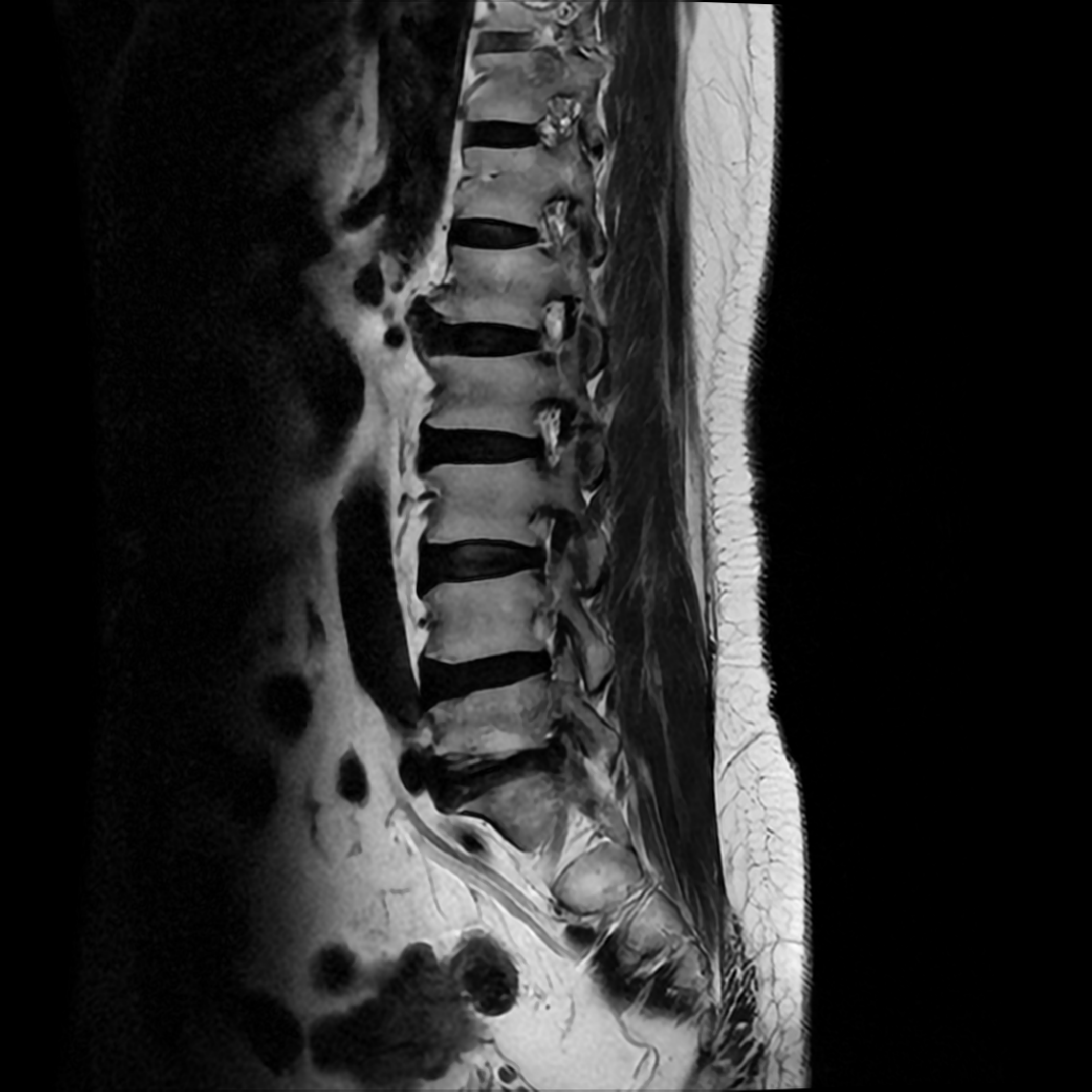}
    }
    \caption{Representative MRI examples (SAG-T2) for the four analyzed categories: Disc Herniation, Disc Bulging, Spondylolisthesis, and Disc Narrowing.}
    \label{fig:pathology_examples}
\end{figure}

The distribution of spinal pathologies within the PhenSPINE dataset exhibits a significant class imbalance, reflecting real-world clinical prevalence where pathological cases are less frequent than normal findings. As presented in Table~\ref{tab:pathology_distribution}, the dataset comprises a total of 1,185 annotated IVD records. Normal cases constitute the majority class, accounting for 66.41\% of the population. Among the specific pathologies, Disc Bulging is the most prevalent condition, occurring in 26.75\% of cases. In contrast, Spondylolisthesis and Disc Narrowing are relatively rare, appearing in only 3.29\% and 4.73\% of records, respectively. Disc Herniation is observed in 6.92\% of the dataset.  Fig.~\ref{fig:pathology_examples} illustrates the distinct morphological characteristics of these four pathology classes, as visualized in the SAG-T2 sequence.

\begin{table}[htbp]
\centering
\caption{Distribution of spinal pathologies in the PhenSPINE dataset}
\label{tab:pathology_distribution}
\begin{tabular}{lcccc}
\toprule
\textbf{Pathology} & \textbf{Pos.} & \textbf{Neg.} & \textbf{Total} & \textbf{Prev. (\%)} \\ \midrule
Herniation & 82 & 1103 & 1185 & 6.92 \\
Bulging & 317 & 868 & 1185 & 26.75 \\
Spondylolisthesis & 39 & 1146 & 1185 & 3.29 \\
Narrowing & 56 & 1129 & 1185 & 4.73 \\
Normal & 787 & 398 & 1185 & 66.41 \\ \bottomrule
\end{tabular}
\end{table}

\subsubsection{Data Preprocessing.}
Missing sequences are handled via zero-padding to preserve consistent input dimensionality. A central slice is selected from each sequence using:
\begin{equation}
Y = \left\lfloor \frac{n}{2} \right\rfloor + 1,
\end{equation}
where $n$ is the total slice count and $Y$ the selected index, ensuring the input corresponds to the anatomical midpoint where pathological features are most pronounced. Raw DICOM images are converted to PNG and resized to $224 \times 224$ pixels. During training, augmentation via \texttt{torchvision} includes geometric transforms (RandomHorizontalFlip, RandomRotation, RandomAffine with translation up to $15\%$ and scale $[0.85, 1.15]$) and photometric transforms (ColorJitter, GaussianBlur $\sigma \in [0.1, 0.5]$ with $p{=}0.3$). Sequence-specific normalization uses adapted ImageNet statistics ($\mu{=}0.449$, $\sigma{=}0.226$) for grayscale inputs. Each IVD is ultimately represented as a multi-sequence tensor $\mathbf{X} \in \mathbb{R}^{4 \times 224 \times 224}$.

\subsection{PhenSPINE Benchmark}
Following the data pre-processing stage to ensure the integrity of the input features, we proposed a pipeline for benchmarking the multi-sequence fusion approach. This framework is specifically designed to evaluate the performance of integrating diverse MRI sequences for detecting complex spinal pathologies.

\begin{figure*}[ht]
    \centering
    \includegraphics[width=\linewidth]{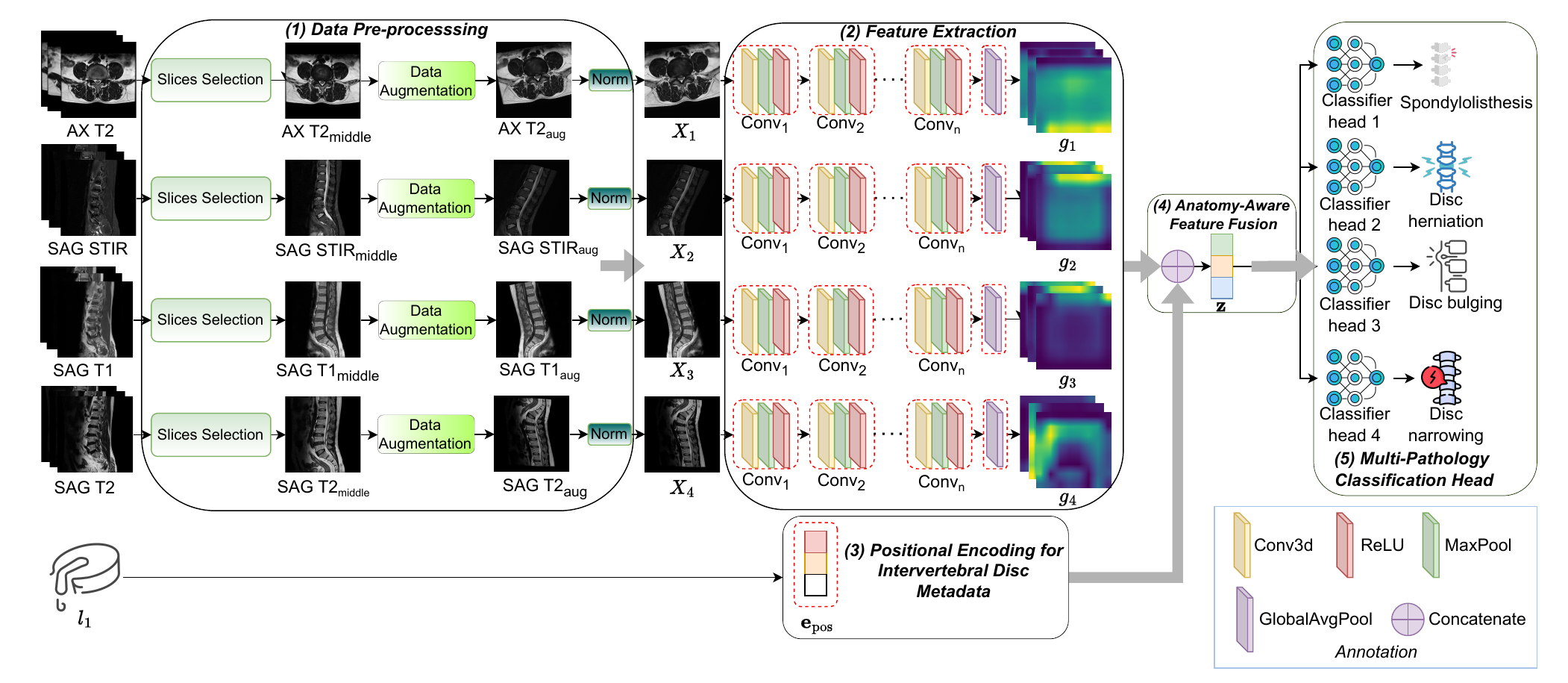}
    \caption{Architecture of the PhenSPINE Benchmark. The pipeline comprises five key stages: (1) Data Pre-processing, (2) Feature Extraction, (3) Positional Encoding for Intervertebral Disc Metadata, (4) Anatomy-Aware Feature Fusion, and (5) Multi-Pathology Classification Head.}
    \label{fig:pipeline}
\end{figure*}

\label{sec:method}
\subsubsection{Feature Extraction.}
Given a collection of multi-sequence MRI scans $\{X_i\}_{i=1}^4$, where each $X_i \in \mathbb{R}^{H \times W \times C}$ corresponds to a specific MRI sequence, high-level spatial feature maps $F_i$ are extracted according to:
\begin{equation}
    F_i = \mathcal{F}(X_i; \theta), \quad i \in \{1, 2, 3, 4\},
\end{equation}
where $\mathcal{F}$ denotes the feature extraction backbone parameterized by $\theta$. 

A global average pooling operation is subsequently applied to each feature map $F_i$ to aggregate local spatial information into a global descriptor:
\begin{equation}
    g_i = \text{GlobalAvgPool}(F_i), \quad i \in \{1, 2, 3, 4\},
\end{equation}
where $g_i \in \mathbb{R}^{C}$ represents the spatially-aggregated feature vector for the $i$-th MRI sequence.

These global descriptors are then directly concatenated along the feature dimension to leverage complementary semantic information across all MRI sequences, forming a unified multi-sequence representation $\mathbf{m}$:
\begin{equation}
    \mathbf{m} = [g_1; g_2; g_3; g_4], \quad \mathbf{m} \in \mathbb{R}^{4C},
\end{equation}
where $C$ specifies the dimensionality of each individual sequence feature vector, and the semicolon notation denotes concatenation.

\subsubsection{Positional Encoding for Intervertebral Disc Metadata.}
Inspired by the Transformer \cite{vaswani2017attention}, we adapted the Positional Encoding mechanism to suit this domain. The categorical metadata specifying the IVD level for each input sequence is mapped to a discrete numerical representation through a label encoding function $E$. Let $\mathcal{L} = \{l_1, l_2, \ldots, l_5\}$ denote the set of lumbar IVD levels, where each level is assigned a unique integer index according to:
\begin{equation}
    y_{\text{ivd}} = E(l_i), \quad y_{\text{ivd}} \in \{0, 1, 2, 3, 4\}.
\end{equation}
The encoded index $y_{\text{ivd}}$ serves as an anatomical identifier that facilitates retrieval of the corresponding positional representation from a learned embedding space.

To effectively encode the distinctive anatomical characteristics associated with each IVD level, we employ a trainable embedding matrix $\mathbf{W}_{\text{pos}} \in \mathbb{R}^{N \times d_s}$, where $N = 5$ represents the total number of lumbar levels and $d_s$ specifies the dimensionality of the positional embedding space. This embedding mechanism projects the discrete level index $y_{\text{ivd}}$ into a continuous, high-dimensional representation according to:
\begin{equation}
    \mathbf{e}_{\text{pos}} = \text{Lookup}(y_{\text{ivd}}, \mathbf{W}_{\text{pos}}), \quad \mathbf{e}_{\text{pos}} \in \mathbb{R}^{d_s},
\end{equation}
where $\mathbf{e}_{\text{pos}}$ denotes the resulting positional embedding vector. This learned representation encodes the spatial characteristics specific to each IVD level, thereby enabling the model to modulate its diagnostic inference according to the anatomical context of the input data.

\subsubsection{Anatomy-Aware Feature Fusion.}
To establish a comprehensive representation integrating both multi-sequence features and anatomical context, the positional embedding $\mathbf{e}_{\text{pos}}$ is combined with the global multi-sequence feature vector $\mathbf{m}$. The anatomy-aware feature vector $\mathbf{z}$ is formulated through concatenation:
\begin{equation}
    \mathbf{z} = [\mathbf{m}; \mathbf{e}_{\text{pos}}], \quad \mathbf{z} \in \mathbb{R}^{4C + d_s},
\end{equation}
where the semicolon denotes concatenation along the feature dimension. This joint representation $\mathbf{z}$ enhances model interpretability and diagnostic robustness across distinct lumbar levels.

\subsubsection{Multi-Pathology Classification Head.}
To detect pathology, the fused anatomy-aware feature vector $\mathbf{z}$ is processed by a multi-head classification architecture comprising four independent feedforward networks. Each classification head $j \in \{1, 2, 3, 4\}$ maps the shared representation $\mathbf{z}$ to a pathology-specific probability $p_j$ through a two-layer nonlinear transformation followed by sigmoid activation:
\begin{equation}
    p_j = \sigma \left( \mathbf{W}_{2,j}^\top \cdot \text{ReLU}(\mathbf{W}_{1,j}^\top \mathbf{z} + \mathbf{b}_{1,j}) + b_{2,j} \right),
\end{equation}
where $\mathbf{W}_{1,j}$, $\mathbf{W}_{2,j}$, $\mathbf{b}_{1,j}$, and $b_{2,j}$ denote the learnable parameters of the $j$-th classification head. The individual probabilities are subsequently assembled into a unified prediction vector $\mathbf{p} = [p_1, p_2, p_3, p_4]^\top$.

\subsubsection{Loss Function.}
We employed the Binary Cross-Entropy loss function for the multi-label classification task, formulated to aggregate the prediction errors across all pathology heads. The objective function is defined as:
\begin{equation}
    \mathcal{L} = - \frac{1}{N} \sum_{i=1}^{N} \sum_{j=1}^{4} \left[ y_{j}^{(i)} \log(p_{j}^{(i)}) + (1 - y_{j}^{(i)}) \log(1 - p_{j}^{(i)}) \right],
\end{equation}
where $N$ is the batch size, $j$ indexes the classification heads, $y_{j}^{(i)}$ denotes the ground truth label, and $p_{j}^{(i)}$ represents the prediction for the $i$-th sample.

\section{Experiments}
\label{sec:experiments}
\subsection{Experimental Setup.}
\label{ssec:exp_setup}
We utilize the PhenSPINE dataset to analyze spinal pathology detection performance. To ensure robust evaluation, the data is chronologically partitioned into training ($70\%$), validation ($15\%$), and testing ($15\%$) subsets. All experiments were implemented using the PyTorch framework on an NVIDIA RTX 4090 GPU with 24~GB of memory. We employ a comprehensive set of metrics including Accuracy, Recall, Precision, Macro F1-score, and AUC to assess model performance. In our experimental tables, the best results are marked in \textbf{bold}, while the second-best results are \underline{underlined}.
\subsection{Comparative Evaluation of Feature Extraction Architectures.}
\begin{table}[ht]
\centering
\caption{Classification performance comparison across backbone architectures}
\label{tab:backbone_comparison}
\begin{tabular}{llccccc}
\toprule
\textbf{Backbone} & \textbf{Sequence} & \textbf{Acc} & \textbf{Recall} & \textbf{Pre} & \textbf{Macro F1} & \textbf{AUC} \\
\midrule
DenseNet \cite{huang2017densely} & AX-T2 & 42.22 & \underline{71.89} & 31.05 & 41.01 & 81.85 \\
& SAG-STIR & 59.44 & 37.15 & 30.81 & 33.14 & 67.28 \\
& SAG-T1 & 57.22 & 50.69 & 38.20 & 40.98 & 82.43 \\
& SAG-T2 & \textbf{66.67} & 54.93 & \underline{48.60} & \underline{49.76} & \underline{88.41} \\
\midrule
ResNet \cite{he2016deep} & AX-T2 & 57.78 & 56.19 & 37.50 & 42.77 & 82.66 \\
& SAG-STIR & 56.67 & 43.89 & 47.49 & 32.35 & 70.08 \\
& SAG-T1 & 52.22 & 54.43 & 31.28 & 37.08 & 79.26 \\
& SAG-T2 & 60.00 & 62.16 & 40.22 & 42.37 & 85.75 \\
\midrule
EffB2 \cite{tan2019efficientnet} & AX-T2 & 56.11 & 67.64 & 38.05 & 46.22 & 85.67 \\
& SAG-STIR & 55.00 & 46.28 & 24.97 & 32.04 & 73.07 \\
& SAG-T1 & 63.89 & 47.14 & 44.70 & 40.22 & 85.67 \\
& SAG-T2 & 54.44 & 64.65 & 43.64 & 47.74 & 87.69 \\
\midrule
EffB1 \cite{tan2019efficientnet} & AX-T2 & 60.56 & 70.11 & 36.33 & 47.16 & 86.50 \\
& SAG-STIR & 56.67 & 40.27 & 25.14 & 30.11 & 74.45 \\
& SAG-T1 & 55.56 & \textbf{71.92} & 33.49 & 43.99 & 85.87 \\
& SAG-T2 & 57.78 & 59.87 & \textbf{60.01} & \textbf{50.31} & 87.46 \\
\midrule
EffB0 \cite{tan2019efficientnet} & AX-T2 & 58.33 & 55.85 & 39.71 & 43.39 & 84.15 \\
& SAG-STIR & 60.00 & 48.60 & 28.01 & 34.68 & 71.02 \\
& SAG-T1 & 63.89 & 48.32 & 36.27 & 40.43 & 86.23 \\
& SAG-T2 & \underline{66.11} & 44.46 & 48.04 & 43.02 & \textbf{88.47} \\
\bottomrule
\end{tabular}
\end{table}
Table 3 presents a comparative evaluation of five backbone architectures across four distinct MRI sequences. The empirical results indicate that the EfficientNet-B1 backbone, when paired with the SAG-T2 sequence, achieves the highest Macro F1-score of 50.31\% and a Precision of 60.01\%, establishing an optimal balance between discriminative power and computational efficiency. While DenseNet with SAG-T2 achieves the highest overall Accuracy (66.67\%) and EfficientNet-B0 with SAG-T2 attains the peak AUC (88.47\%), EfficientNet-B1 consistently demonstrates superior F1-scores . Conversely, larger architectures like EfficientNet-B2 and ResNet fail to surpass the performance of the more compact B1 variant, suggesting that increased model complexity does not necessarily yield enhanced diagnostic accuracy in this domain.

From this evaluation, we derive a primary observation: EfficientNet-B1 demonstrates a superior ability to mitigate the overfitting often observed in deeper networks while retaining sufficient capacity to capture subtle pathological features. Consequently, we select the EfficientNet-B1 architecture as the primary backbone for subsequent multi-sequence fusion experiments.

\subsection{Evaluation of Single-Sequence Models without IVD Metadata.}

\begin{table}[htbp]
\centering
\caption{Diagnostic performance across different MRI sequences without IVD metadata}
\label{tab:results_no_ivd}
\begin{tabular}{lrrrrr}
\toprule
\textbf{Sequence} & \textbf{Acc} & \textbf{Recall} & \textbf{Pre} & \textbf{F1} & \textbf{AUC} \\
\midrule
SAG-T1 & 2.78 & \textbf{60.81} & 23.06 & 26.96 & \underline{66.13} \\
SAG-T2 & \underline{36.11} & \underline{49.67} & \textbf{28.12} & \textbf{27.92} & \textbf{73.61} \\
AX-T2 & \textbf{42.22} & 33.44 & \underline{27.27} & \underline{27.23} & 64.91 \\
SAG-STIR & 6.11 & 43.30 & 11.91 & 15.84 & 56.13 \\
\bottomrule
\end{tabular}
\end{table}
Table~\ref{tab:results_no_ivd} details the diagnostic performance of backbone models when trained on individual MRI sequences without the inclusion of IVD metadata. The empirical evidence reveals that while T2-weighted modalities generally exhibit superior performance metrics-with SAG-T2 attaining the highest F1-score (27.92\%) and AX-T2 reporting the highest Accuracy (42.22\%)-all sequences struggle notably compared to models integrated with IVD metadata. These results underscore a fundamental characteristic of this diagnostic task: the pronounced performance degradation across all sequences highlights that IVD positional metadata is not merely an auxiliary feature but a prerequisite for enabling the model to localize and differentiate vertebral structures accurately.

\subsection{Impact of Positional Encoding for Intervertebral Disc Metadata.}
\label{ssec:ivd_encoding_impact}
\begin{table}[ht]
\centering
\caption{Comparison of IVD encoding methods}
\label{tab:ivd_encoding_comparison}
\begin{tabular}{lcccccccccc}
\toprule
& \multicolumn{5}{c}{\textbf{Label Encoding}} & \multicolumn{5}{c}{\textbf{Positional Encoding}} \\
\cmidrule(lr){2-6} \cmidrule(lr){7-11}
\textbf{Sequence} & \textbf{Acc} & \textbf{Recall} & \textbf{Pre} & \textbf{F1} & \textbf{AUC} & \textbf{Acc} & \textbf{Recall} & \textbf{Pre} & \textbf{F1} & \textbf{AUC} \\
\midrule
SAG-T1 & \underline{63.33} & 58.87 & 37.72 & 44.61 & 86.87 & 55.56 & \textbf{71.92} & 33.49 & 43.99 & 85.87 \\
SAG-T2 & \textbf{68.33} & \textbf{62.85} & \underline{39.58} & \textbf{47.60} & \textbf{87.56} & \underline{57.78} & 59.87 & \textbf{60.01} & \textbf{50.31} & \textbf{87.46} \\
AX-T2  & 58.89 & \underline{60.83} & \textbf{40.90} & \underline{44.99} & \underline{87.40} & \textbf{60.56} & \underline{70.11} & \underline{36.33} & \underline{47.16} & \underline{86.50} \\
SAG-STIR & 58.33 & 37.44 & 29.53 & 32.91 & 75.52 & 56.67 & 40.27 & 25.14 & 30.11 & 74.45 \\
\bottomrule
\end{tabular}
\end{table} 
The comparative results presented in Table~\ref{tab:ivd_encoding_comparison} quantify the significant contribution of Positional Encoding for IVD metadata. Specifically, for the optimal SAG-T2 sequence, while standard Label Encoding yields a higher raw Accuracy of $68.33\%$, it is compromised by a substantially lower Precision ($39.58\%$) and Macro F1-score ($47.60\%$). In contrast, the integration of Positional Encoding elevates Precision to $60.01\%$ and the F1-score to $50.31\%$. These empirical observations underscore the efficacy of Positional Encoding as a spatial inductive bias essential for spinal pathology localization. Positional Encoding maps IVD levels into a continuous vector space, thereby capturing relative spatial distances and sequential dependencies between levels. This structure-aware representation effectively guides the backbone network to differentiate between similar visual patterns at distinct anatomical locations, transitioning the diagnostic process from image classification toward a more context-aware clinical assessment.

\subsection{Multi-Sequence Fusion Strategies.}
\begin{table}[ht]
\centering
\caption{Classification performance on different sequence MRI combinations}
\label{tab:sequence_combinations}
\begin{tabular}{lrrrrr}
\toprule
\textbf{Methods} & \textbf{Acc} & \textbf{Recall} & \textbf{Pre} & \textbf{F1} & \textbf{Auc} \\
\midrule
SAG-T1 & 55.56 & \underline{71.92} & 33.49 & 43.99 & 85.87 \\
SAG-T2 & 57.78 & 59.87 & \textbf{60.01} & \textbf{50.31} & 87.46 \\
SAG-STIR & 56.67 & 40.27 & 25.14 & 30.11 & 74.45 \\
AX-T2 & 60.56 & 70.11 & 36.33 & 47.16 & 86.50 \\
SAG-T1+AX-T2 & 56.11 & 48.16 & 40.73 & 42.05 & 78.24 \\
SAG-T1+SAG-T2 & 64.44 & 67.48 & 39.42 & 48.74 & 88.84 \\
SAG-T1+SAG-STIR & 63.33 & \textbf{73.98} & 38.35 & \underline{49.44} & \textbf{89.30} \\
SAG-T2+AX-T2 & 55.00 & 51.29 & \underline{52.06} & 44.29 & 87.47 \\
SAG-T2+SAG-STIR & 54.44 & 58.70 & 38.88 & 44.77 & 86.65 \\
AX-T2+SAG-STIR & 58.89 & 54.27 & 30.65 & 37.83 & 84.69 \\
SAG-T1+SAG-T2+AX-T2 & \underline{65.00} & 49.51 & 45.19 & 45.27 & \underline{88.94} \\
SAG-T1+SAG-T2+SAG-STIR & 63.89 & 52.16 & 42.32 & 45.96 & 87.03 \\
SAG-T1+AX-T2+SAG-STIR & 58.89 & 56.77 & 41.09 & 45.74 & 85.60 \\
SAG-T2+AX-T2+SAG-STIR & 58.89 & 60.35 & 39.69 & 45.07 & 86.73 \\
All Sequences & \textbf{67.22} & 61.63 & 40.37 & 48.33 & 88.86 \\
\bottomrule
\end{tabular}
\end{table}

Table~\ref{tab:sequence_combinations} presents the classification performance across various MRI sequence combinations. The SAG-T2 sequence demonstrates superior performance across critical discriminative metrics, enabling an efficient diagnostic model. Specifically, SAG-T2 achieves a Macro F1-score of $50.31\%$ and a Precision of $60.01\%$, substantially outperforming the multi-sequence fusion models and other single sequences. Notably, while the All Sequences fusion strategy achieves the highest global Accuracy ($67.22\%$) and the combination of SAG-T1 and SAG-STIR yields the peak AUC ($89.30\%$), these gains do not translate to better pathology-specific detection as evidenced by lower F1-scores ($48.33\%$ and $49.44\%$ respectively).

The comparative analysis highlights the specialized efficacy of single-sequence feature learning for spinal diagnostics. Although multi-sequence integration enhances broad performance metrics such as global Accuracy and AUC, the SAG-T2 sequence possesses high information density, enabling the network to distill essential diagnostic markers without noise introduced by auxiliary sequences. Therefore, the combination of SAG-T2 and EfficientNet-B1 is established as the optimal configuration, offering a clinically interpretable benchmark for characterizing spinal pathologies.

\section{Conclusion}
\label{sec:conclusion}
In this study, we introduced PhenSPINE, a comprehensive MRI dataset designed to advance spinal pathology diagnosis. By curating a diverse dataset and establishing a rigorous evaluation protocol, we analyzed the efficacy of different MRI sequences and deep learning architectures. Our benchmark, which integrates a Positional Encoding mechanism, demonstrated the critical importance of anatomical context, significantly outperforming standard label encoding approaches. Extensive experiments revealed that the SAG-T2-weighted sequence provides the most discriminative features for pathology detection, whereas multi-sequence fusion strategies failed to yield performance gains due to feature redundancy and noise in our dataset. To address these limitations, future work will focus on incorporating explicit spine segmentation into the pipeline to precisely isolate vertebral structures, thereby eliminating background noise and facilitating more effective multi-sequence integration. Finally, to support the biomedical research community and foster further innovation, we are committed to making the PhenSPINE dataset publicly available on an appropriate open-access platform.

\section*{Acknowledgement}
This research is supported by the Ben Dam Me Award Fund, the Vietnam Young Talent Support Fund, and the Number One Brand, Tan Hiep Phat Group.

\bibliographystyle{splncs04}
\bibliography{references}
\end{document}